%% file: main.tex
\documentclass[10pt,twocolumn,letterpaper]{article}
\usepackage[pagenumbers]{cvpr}
\input{preamble}

\usepackage{longtable}
\definecolor{cvprblue}{rgb}{0.21,0.49,0.74}
\usepackage[pagebackref,breaklinks,colorlinks,allcolors=cvprblue]{hyperref}
\usepackage[capitalize]{cleveref}
\hypersetup{pdftitle={JEPA Dots: Dual-Latent World Model for Language-Guided Robot Control},pdfauthor={Xianhao Pan, Chi Li, Hanchao Li, Yunxin Liu, Jiankai Sun, Luhui Hu}}
\title{JEPA Dots: Dual-Latent World Model\\for Language-Guided Robot Control}
\input{authors}
\begin{document}
% Keep natural float/text spacing instead of stretching gaps to align column bottoms.
\raggedbottom
\maketitle
\input{sections/00_abstract}
\input{sections/01_introduction}
\input{sections/02_related}
\input{sections/03_method}
\input{sections/04_experiments}
\input{sections/05_results}
\input{sections/06_discussion}
\label{sec:main_end}
\input{sections/07_llm_statement}
{\small\bibliographystyle{ieeenat_fullname}\bibliography{references,references_extra}}

% The complete v13 supplement follows the references, with unchanged content.
\clearpage
\onecolumn
\setcounter{section}{0}
\setcounter{subsection}{0}
\setcounter{table}{0}
\setcounter{figure}{0}
\renewcommand{\thesection}{S\arabic{section}}
\renewcommand{\thetable}{S\arabic{table}}
\renewcommand{\thefigure}{S\arabic{figure}}
\renewcommand{\theHsection}{supp.\arabic{section}}
\renewcommand{\theHtable}{supp.\arabic{table}}
\renewcommand{\theHfigure}{supp.\arabic{figure}}
\makeatletter\setlength{\@fptop}{0pt}\makeatother
\phantomsection
\pdfbookmark[0]{Supplementary Material}{supplementary.material}
\label{sec:supplement_start}
\begin{center}\Large\bfseries Supplementary Material\end{center}
\vspace{6pt}
\input{sections/supplementary_body}

\end{document}

%% file: preamble.tex
\usepackage{graphicx}
\usepackage{amsmath,amssymb}
\usepackage{booktabs,multirow,array}
\usepackage{dcolumn}
\newcolumntype{d}[1]{D{.}{.}{#1}}
\usepackage{xspace}
\newcommand{\method}{JEPA Dots\xspace}

\newcommand{\hist}{\mathcal{H}_t}
\newcommand{\goals}{\mathcal{G}_t}
\newcommand{\E}{\mathbb{E}}
\newcommand{\R}{\mathbb{R}}

%% file: authors.tex
\author{%
Xianhao~Pan$^{1}$\quad Chi~Li$^{1}$\quad Hanchao~Li$^{1}$\quad
Yunxin~Liu$^{2}$\quad Jiankai~Sun$^{3}$\quad Luhui~Hu$^{1}$\thanks{Corresponding author.}\\[4pt]
$^{1}$Z Robotics\quad $^{2}$Tsinghua University\quad $^{3}$Stanford University
}

%% file: sections/00_abstract.tex
\begin{abstract}
Predictive world models for language-guided robot control face a fundamental conditioning challenge: instructions should influence action selection without altering the predicted consequences of fixed actions under identical observations. To address this challenge, we introduce JEPA Dots (Dynamics-Oriented Task Separation), a compact dual-latent framework that jointly learns visual dynamics and language-guided control through separate dynamics and task representations. Language-conditioned task features guide a flow-matching policy, while masked attention restricts world prediction to instruction-independent dynamics history and supplied actions. In closed-loop evaluation on 40 trained LIBERO tasks, JEPA Dots achieves 84.9\% overall success with approximately 39.1M parameters, compared with the published 78.7\% of NanoVLA-S with 161.0M parameters under a different evaluation protocol. In the reference experiment, JEPA Dots also improves overall success from 76.0\% for our task-ID-conditioned adaptation baseline to 84.9\%, with LIBERO-Goal success increasing from 67.1\% to 88.5\%. Instruction-switching experiments further confirm that task representations and policy outputs respond to language changes, while dynamics representations and world predictions remain invariant under fixed observations and supplied actions.
\end{abstract}

%% file: sections/01_introduction.tex
\section{Introduction}\label{sec:introduction}
\begin{figure*}[t]
\centering
\includegraphics[width=\textwidth]{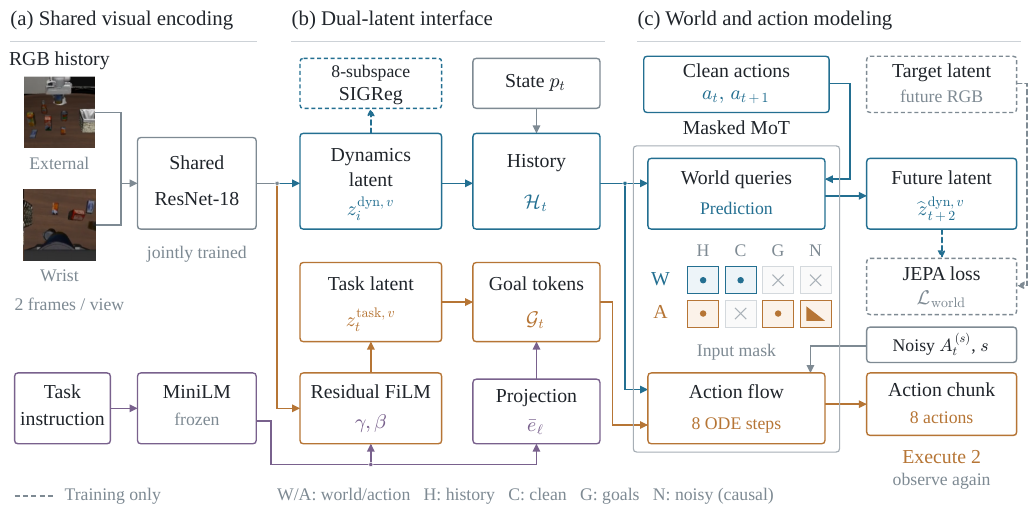}
\caption{JEPA Dots separates instruction-independent dynamics representations from language-conditioned task representations within a shared visual architecture. Masked world queries read only history and supplied clean actions; the policy reads history, task goals, and noisy actions. Future-latent targets and subspace regularization supervise training, while a flow-matching policy generates action chunks for receding-horizon control.}
\label{fig:architecture}
\end{figure*}

Joint-embedding predictive architectures (JEPAs)~\cite{assran2023ijepa,bardes2024vjepa} have emerged as a promising approach to visual representation learning by predicting future or missing information in latent space. Their ability to learn compact visual representations and predictive dynamics offers considerable potential for downstream prediction and planning~\cite{assran2025vjepa2,murlabadia2026vjepa21,zhang2026hjepa,zhou2025dinowm}. Recent studies have further extended JEPA-based learning to language-guided robot control, including vision--language--action modeling~\cite{lin2026jepawam,sun2026vlajepa}, stage-level prediction~\cite{liu2026stagejepa}, and compact world--action policies~\cite{clemente2026lajepa}.

However, integrating language conditioning into JEPA-based world models poses a fundamental challenge: task instructions should influence action selection without altering the predicted consequences of fixed actions under identical observations. When language conditioning is applied to shared visual representations, task-dependent information may also influence dynamics prediction, making it difficult to maintain this separation.

Recent advances in JEPA-based world modeling have explored subspace regularization~\cite{zhao2026subjepa}, temporal centering~\cite{liu2026tclewm}, structured representations~\cite{huang2026lrcjepa}, flow-based dynamics~\cite{huo2026flowjepa}, and hierarchical or amortized planning~\cite{bansal2026leplanner,huang2026leflow,zhang2026hjepa}. LeWorldModel (LeWM)~\cite{maes2026lewm} jointly learns visual representations and action-conditioned dynamics directly from pixels, while LeWAM~\cite{fu2026lewam} extends this formulation to joint world--action modeling through masked attention. Building on LeWM, TC-LeWM~\cite{liu2026tclewm} applies Gaussian regularization to temporally centered residuals to reduce the dominance of persistent scene information and better preserve dynamics-relevant variations. Despite these advances in predictive representation learning and world--action modeling, explicitly separating language-conditioned task information from instruction-independent dynamics prediction remains an important architectural consideration.

We propose JEPA Dots (Dynamics-Oriented Task Separation), a compact framework that explicitly separates language-conditioned task representations from instruction-independent dynamics representations within a jointly trained world--action model (\cref{fig:architecture}). Building on LeWM's end-to-end predictive learning~\cite{maes2026lewm} and LeWAM's masked world--action modeling~\cite{fu2026lewam}, JEPA Dots employs a shared ResNet-18 backbone~\cite{he2016resnet} with two distinct latent pathways. The dynamics pathway extracts visual representations without language conditioning, while the task pathway applies language-conditioned residual FiLM~\cite{perez2018film} to produce task-specific representations. These task representations are combined with language embeddings to form goal tokens, which condition a flow-matching policy~\cite{lipman2022flow} for action generation.

To preserve this separation, masked attention restricts world prediction to dynamics history and a prefix of clean actions, while the policy attends to dynamics history, goal tokens, and noisy actions. This design ensures that world predictions remain invariant to language changes when model parameters, observations, and supplied actions are fixed, while allowing task representations and policy outputs to adapt to different instructions. We further apply eight-subspace Gaussian regularization~\cite{zhao2026subjepa} to the dynamics representations during training. The visual and control modules are jointly optimized through predictive and action objectives, resulting in a compact network with 16.4M parameters, or approximately 39.1M including the frozen language encoder.

We evaluate JEPA Dots on 40 trained tasks across four LIBERO suites~\cite{liu2023libero}. In the reference experiment, JEPA Dots achieves an overall closed-loop success rate of 84.9\% with approximately 39.1M parameters, compared with the published results of NanoVLA-S (78.7\%, 161.0M parameters)~\cite{chen2025nanovla} and SmolVLA-240M (82.8\%, 240.0M parameters)~\cite{shukor2025smolvla}, although their training and evaluation protocols differ. Under our evaluation protocol, JEPA Dots further improves overall success from 76.0\% for the task-ID-conditioned adaptation baseline to 84.9\%, with LIBERO-Goal success increasing from 67.1\% to 88.5\%. Instruction-switching experiments demonstrate that task representations and policy outputs respond to language changes, while dynamics representations and world predictions remain unchanged under fixed observations and supplied actions. Additional matched ablations examine the effects of language-conditioned feature modulation and subspace Gaussian regularization on control performance and latent prediction.

The main contributions of this work are summarized as follows:
\begin{itemize}
\item We design a dual-representation interface for joint world--action modeling, providing distinct conditioning paths for instruction-dependent action generation and dynamics prediction without direct instruction input.
\item We integrate this interface with JEPA future-latent prediction, a flow-matching policy, and subspace Gaussian regularization, using attention masks to enforce prediction invariance for fixed observations and supplied actions.
\item We analyze closed-loop control, instruction switching, and component ablations on 40 trained LIBERO tasks, reporting variation across training runs and task-dependent FiLM effects.
\end{itemize}

%% file: sections/02_related.tex
\section{Related work}\label{sec:related}
\paragraph{Joint world--action learning and language conditioning.}
Recent JEPA-based approaches have integrated predictive learning with robot control. LeWAM~\cite{fu2026lewam} combines latent world prediction and flow-matching action generation through masked attention. VLA-JEPA~\cite{sun2026vlajepa} incorporates latent prediction into vision--language--action learning, while JEPA-WAM variants explore stage-level prediction~\cite{liu2026stagejepa} and structured transition modeling~\cite{lin2026jepawam}. ThinkJEPA~\cite{zhang2026thinkjepa} incorporates vision--language reasoning features into predictive learning, and La-JEPA~\cite{clemente2026lajepa} develops compact world--action policies using pretrained V-JEPA representations. Despite these advances, the separation of language-conditioned task information from action-conditioned dynamics prediction remains insufficiently explored in jointly learned world--action models.

\paragraph{Predictive representations and dynamics regularization.}
I-JEPA~\cite{assran2023ijepa} and V-JEPA~\cite{bardes2024vjepa} establish latent predictive learning for images and videos, while V-JEPA 2 and 2.1~\cite{assran2025vjepa2,murlabadia2026vjepa21} extend pretrained representations toward downstream applications. DINO-WM~\cite{zhou2025dinowm} models dynamics over pretrained visual features~\cite{oquab2024dinov2}, whereas LeWorldModel (LeWM)~\cite{maes2026lewm} jointly learns visual representations and action-conditioned dynamics directly from pixels. Representation regularization has been explored through VICReg~\cite{bardes2022vicreg}, LeJEPA~\cite{balestriero2025lejepa}, Sub-JEPA~\cite{zhao2026subjepa}, and temporally centered SIGReg in TC-LeWM~\cite{liu2026tclewm}. Other studies investigate dynamics--context separation~\cite{huang2026lrcjepa}, flow-based dynamics~\cite{huo2026flowjepa}, and hierarchical or amortized planning~\cite{bansal2026leplanner,huang2026leflow,zhang2026hjepa}. However, these methods primarily focus on representation quality and predictive modeling, leaving the distinct conditioning requirements of task-dependent control and world prediction less explored.

\paragraph{Language-conditioned policies and compact control.}
FiLM~\cite{perez2018film} enables language-conditioned feature modulation, while BAKU~\cite{haldar2024baku} incorporates task-conditioned visual representations into multitask policies. MiniLM and Sentence-BERT provide compact language embeddings~\cite{reimers2019sbert,sentence2026minilm,wang2020minilm}. Action generation has been studied through action chunking~\cite{zhao2023act}, diffusion policies~\cite{chi2023diffusion}, and flow matching~\cite{lipman2022flow}. Recent methods, including NanoVLA~\cite{chen2025nanovla}, SmolVLA~\cite{shukor2025smolvla}, VLA-Adapter~\cite{wang2025vlaadapter}, and SimVLA~\cite{luo2026simvla}, further explore compact or efficient robot policies. While these approaches advance language-conditioned action generation and model efficiency, they do not directly establish how task conditioning should be separated from dynamics prediction in a shared world--action architecture.

%% file: sections/03_method.tex
\section{Dynamics-Oriented Task Separation}
\label{sec:method}

JEPA Dots jointly learns visual dynamics prediction and language-guided action generation while explicitly separating their conditioning pathways. As illustrated in \cref{fig:architecture}, the framework employs a shared visual encoder with two distinct latent representations to accommodate the different requirements of world prediction and task-dependent control. Language information is incorporated into task representation learning and policy conditioning, whereas dynamics prediction depends only on visual history and supplied actions. This design enables task-specific action generation while maintaining instruction-invariant world prediction under fixed observations and actions. The shared visual backbone is jointly optimized through predictive and action objectives.

\subsection{Dual-latent representation and language conditioning}
\label{sec:dual_latent}
At time $t$, the model receives two-frame RGB observations $o_{t-1:t}^{v}$ from an external camera and a wrist camera, where $v\in\{1,2\}$, together with proprioceptive state $p_t\in\R^{15}$ and language instruction $\ell$. A shared ResNet-18 backbone $B_\theta$ extracts visual feature maps $F_i^v=B_\theta(o_i^v)$. The instruction is encoded by a frozen \texttt{all-MiniLM-L6-v2} encoder~\cite{sentence2026minilm} into $e_\ell\in\R^{384}$, which is further mapped to a 192-dimensional embedding $\bar e_\ell$ through a trainable projection.

\paragraph{Dynamics representation.}
The dynamics pathway extracts visual features without language conditioning. Specifically, a $1\times1$ convolution, spatial softmax~\cite{finn2015spatial}, and an MLP project each feature map into a 192-dimensional latent:
\begin{equation}
z_i^{\mathrm{dyn},v}=P_{\mathrm{dyn}}\bigl(\operatorname{SS}(C_{\mathrm{dyn}}(F_i^v))\bigr)\in\R^{192}.
\label{eq:dynamics_latent}
\end{equation}
The dynamics latents from both camera views and history frames are combined with an embedding of $p_t$ to form the dynamics context $\hist$, which provides shared visual and proprioceptive information for world prediction and action generation.

\paragraph{Task representation.}
Unlike the dynamics pathway, the task pathway incorporates instruction information through residual Feature-wise Linear Modulation (FiLM)~\cite{perez2018film} before spatial pooling:
\begin{equation}
\widetilde F_i^v(\ell)=F_i^v+\gamma_\omega(e_\ell)\odot F_i^v+\beta_\omega(e_\ell).
\label{eq:film}
\end{equation}
The language-conditioned feature maps are then projected into task-specific latents:
\begin{equation}
z_i^{\mathrm{task},v}(\ell)=P_{\mathrm{task}}\bigl(\operatorname{SS}(C_{\mathrm{task}}(\widetilde F_i^v(\ell)))\bigr).
\label{eq:task_latent}
\end{equation}
Here, $\gamma_\omega$ and $\beta_\omega$ generate channel-wise scaling and shifting parameters, respectively. Both are initialized to zero, making residual FiLM an identity mapping at initialization. The task and dynamics heads have separate parameters and each produces 192-dimensional representations per camera view.

Finally, the current task latents are fused with the projected language embedding to construct one goal token per camera:
\begin{equation}
g_t^v=G_\eta(z_t^{\mathrm{task},v},\bar e_\ell),\qquad\goals=\{g_t^1,g_t^2\}.
\label{eq:goal}
\end{equation}
These goal tokens provide language-conditioned visual information for subsequent action generation, while the dynamics representations remain independent of instruction inputs in the forward computation.

\subsection{Masked world--action modeling}
\label{sec:mask}
Given the dynamics context $\hist$ and language-conditioned goal tokens $\goals$ defined in \cref{sec:dual_latent}, we integrate world prediction and action generation through a masked mixture-of-transformers (MoT), following LeWAM~\cite{fu2026lewam}. The MoT consists of four layers with four attention heads and a hidden dimension of 192, employing modality-specific processing and attention masks~\cite{vaswani2017attention} to control information flow between the two branches. Although both branches operate within a jointly trained architecture, they receive distinct conditioning inputs.

\paragraph{World prediction.}
The world branch predicts future dynamics latents from the observation history and a two-step prefix of clean actions:
\begin{equation}
\widehat z^{\mathrm{dyn},v}_{t+2}=f_{\psi,v}(\hist,a_t,a_{t+1}).
\label{eq:world}
\end{equation}
World-state queries attend only to dynamics history and supplied clean actions, without access to language-conditioned goal tokens or noisy policy actions. This restriction prevents task-conditioning signals from entering the world prediction path.

\paragraph{Action generation.}
In contrast, the policy branch generates actions using flow matching~\cite{lipman2022flow}, conditioned on dynamics history, task-specific goal tokens, noisy action inputs, and flow time $s$:
\begin{equation}
\widehat V_t=v_\xi(A_t^{(s)},s;\hist,\goals).
\label{eq:velocity}
\end{equation}
The policy attends to noisy action tokens causally and cannot access the clean actions supplied to the world branch. The attention constraints also cover intermediate tokens, preventing indirect information exchange between the two conditioning pathways. The permitted inputs for both branches are summarized in \cref{tab:access}.

\begin{table}[t]
\centering\small
\caption{\textbf{Inputs to the two branches.} World prediction reads dynamics history and clean actions. Goal tokens and noisy actions condition the policy.}
\label{tab:access}
\setlength{\tabcolsep}{3pt}
\begin{tabular}{@{}lcccc@{}}\toprule
Output & History & Clean & Goals & Noisy\\
& $\hist$ & actions & $\goals$ & actions\\\midrule
World prediction & Yes & Prefix & No & No\\
Action velocity & Yes & No & Yes & Causal\\\bottomrule
\end{tabular}
\end{table}

\paragraph{Instruction-invariant prediction.}
The separation of conditioning inputs ensures that world prediction does not explicitly depend on language instructions. For fixed model parameters, observations, proprioception, and supplied actions, the dynamics context $\hist$ remains unchanged across different instructions. Consequently,
\begin{equation}
\widehat Z_{t+2}(\hist,A;\ell_1)=\widehat Z_{t+2}(\hist,A;\ell_2).
\label{eq:invariance}
\end{equation}
This property applies to predictions conditioned on identical observations and supplied actions. The policy, however, may generate different actions under different instructions, resulting in different future trajectories. Importantly, the separation is imposed on the conditioning pathways rather than requiring statistical independence between the learned representations, since both pathways share a jointly optimized visual backbone.

\subsection{Learning objectives}
JEPA Dots is jointly optimized using latent dynamics prediction, action flow matching, and subspace Gaussian regularization.

\paragraph{World prediction loss.}
Following the JEPA predictive learning formulation, the world branch minimizes the mean squared error between predicted future dynamics latents and the target representations extracted from future observations, with $d=192$:
\begin{equation}
\mathcal L_{\mathrm{world}}=\E\!\left[\frac{1}{2d}\sum_{v=1}^2\|\widehat z^{\mathrm{dyn},v}_{t+2}-z^{\mathrm{dyn},v}_{t+2}\|_2^2\right].
\label{eq:world_loss}
\end{equation}
Future observations are used only to construct prediction targets during training.

\paragraph{Action flow-matching loss.}
Given a demonstration action chunk $A_t\in\R^{8\times7}$, we sample Gaussian noise $\epsilon\sim\mathcal N(0,I)$ and flow time $s\sim\mathcal U(0,1)$. The interpolated action is defined as
\begin{equation}
A_t^{(s)}=(1-s)\epsilon+sA_t.
\label{eq:flow_path}
\end{equation}
The policy learns to predict the corresponding velocity field through
\begin{equation}
\mathcal L_{\mathrm{act}}=\E\bigl[\|v_\xi(A_t^{(s)},s;\hist,\goals)-(A_t-\epsilon)\|^2\bigr].
\label{eq:act_loss}
\end{equation}
The loss is averaged over valid action coordinates, with padded positions at demonstration boundaries excluded.

\paragraph{Subspace Gaussian regularization.}
To regularize the dynamics representations, we adopt the subspace Gaussian regularization proposed in Sub-JEPA~\cite{zhao2026subjepa}. The 192-dimensional dynamics latent is projected onto eight fixed 24-dimensional subspaces, each defined by an orthonormal basis $U_k\in\R^{192\times24}$. The regularization objective is
\begin{equation}
\mathcal L_{\mathrm{sub}}=\frac{1}{8}\sum_{k=1}^{8}\operatorname{SIGReg}(ZU_k),\quad U_k^\top U_k=I_{24}.
\label{eq:k8}
\end{equation}
The subspace bases remain fixed throughout training, with a total budget of 1,024 random projections. The overall training objective is
\begin{equation}
\mathcal L=\mathcal L_{\mathrm{act}}+\mathcal L_{\mathrm{world}}+\ensuremath{4.0\times10^{-2}}\mathcal L_{\mathrm{reg}},
\label{eq:total}
\end{equation}
where $\mathcal L_{\mathrm{reg}}=\mathcal L_{\mathrm{sub}}$ for the complete JEPA Dots model. The predictive and action objectives jointly update the shared visual backbone and their respective modules, while Gaussian regularization constrains the dynamics representations during training without introducing additional inference-time computation.

\subsection{Compact closed-loop control}
During inference, JEPA Dots takes multi-view visual observations, proprioceptive states, and language instructions as inputs. The shared visual encoder produces dynamics and task representations, which are combined with language embeddings to condition the flow-matching policy. At each control step, the policy generates an eight-action chunk using eight ODE integration steps and executes the first two actions before acquiring new observations. This receding-horizon procedure enables closed-loop control without relying on world-model planning, as the prediction branch is used only for training supervision.

The visual and control network contains 16.4M parameters, including 197,632 parameters introduced by residual FiLM. With the frozen sentence encoder contributing approximately 22.7M parameters, the complete model contains approximately 39.1M parameters, providing a compact architecture for jointly learned visual dynamics and language-guided control.

%% file: sections/04_experiments.tex
\section{Experiments and results}\label{sec:experiments}\label{sec:results}
\subsection{Experimental setup}
\paragraph{Tasks and datasets.}
We evaluate JEPA Dots on 40 tasks from four LIBERO suites~\cite{liu2023libero}: Spatial, Object, Goal, and LIBERO-10, with ten tasks per suite. Language instructions are obtained from the demonstrations, and numeric task identifiers are excluded from language-conditioned inputs. We reserve 10.0\% of demonstrations for validation at the trajectory level.

\paragraph{Training configuration.}
The model is trained for 60 epochs using AdamW with a learning rate of $1.0\times10^{-4}$, an effective batch size of 64, and approximately 15,000 optimizer updates. Training jointly optimizes action flow matching, latent world prediction, and subspace Gaussian regularization.

\paragraph{Baselines and ablations.}
We compare JEPA Dots with frozen V-JEPA 2.1-B/L encoders coupled with flow matching, as well as published NanoVLA-S~\cite{chen2025nanovla} and SmolVLA-240M~\cite{shukor2025smolvla} results. The V-JEPA baselines share the action-flow architecture and update budget with JEPA Dots but differ in encoder training, prediction objectives, and preprocessing. We additionally evaluate a task-ID-conditioned reference and matched ablations of the task-visual branch, residual FiLM, and subspace regularization. Published VLA results retain their original evaluation protocols.

\paragraph{Evaluation protocol.}
Closed-loop evaluation uses 50 official initial states per task under two simulation random streams, totaling 4,000 episodes per model in the reference experiment. Each episode is limited to 600 environment steps. The policy predicts eight-step action chunks and executes the first two actions before replanning. To assess training variability, we additionally report three independent training runs for the main ablations, with 12,000 evaluation episodes per variant. The reference result and repeated-run statistics are reported separately. Further implementation details, baseline configurations, and evaluation protocols are provided in the supplementary material. Differences and relative changes are computed from unrounded records; displayed measurements are rounded to one decimal.

%% file: sections/05_results.tex
\subsection{Comparison with other models}\label{sec:compact}
\input{data/external_table}
\Cref{tab:external} compares JEPA Dots with V-JEPA 2.1-based flow-matching baselines and representative compact VLA models. Compared with V-JEPA 2.1-B + FM and V-JEPA 2.1-L + FM, which achieve overall success rates of 68.8\% and 69.9\%, respectively, JEPA Dots achieves 84.9\%, improving performance by approximately 16.1 and 15.0 percentage points. Notably, simply increasing the pretrained visual encoder size from ViT-B to ViT-L yields only marginal improvement. These results suggest that coupling pretrained JEPA representations with a flow-matching policy alone may be insufficient for effective robotic control. In contrast, our dual-latent architecture jointly learns action-conditioned dynamics representations and language-conditioned task representations, providing a more structured interface between predictive representation learning and action generation.

Compared with recent compact VLA models, JEPA Dots also achieves competitive performance with substantially fewer parameters. Specifically, it outperforms NanoVLA-S (78.7\%, 161.0M parameters) and SmolVLA-240M (82.8\%, 240.0M parameters) in overall success rate, while requiring only approximately 39.1M parameters. The advantage is particularly evident on LIBERO-10, where JEPA Dots achieves 64.8\% success, compared with 49.8\% and 63.0\% for NanoVLA-S and SmolVLA-240M, respectively. Although these published results follow different evaluation protocols, the comparison highlights the potential of compact JEPA-based architectures to achieve strong manipulation performance without relying on substantially larger VLA models.

The results suggest that effective action control depends not only on model capacity but also on how visual representations are organized and optimized for downstream actions. By separating instruction-independent dynamics prediction from language-conditioned task representations, JEPA Dots enables jointly learned JEPA features to support flow-matching action generation within a compact architecture. This provides evidence that structured latent learning is a promising alternative to simply scaling model parameters for language-guided robotic manipulation.

\begin{figure}[t]
\centering\includegraphics[width=\columnwidth]{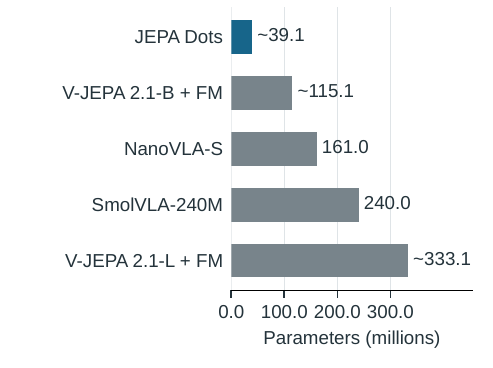}
\caption{\textbf{Total parameter counts.} DOTS and V-JEPA 2.1 + flow matching include frozen MiniLM (approximately 22.7M). Published policy counts follow their sources. Parameter count does not measure latency.}
\label{fig:parameters}
\end{figure}

\subsection{Effect of task-aware representations}
\input{data/reference_table}
We first investigate the contribution of task-aware representations to language-guided control. As shown in \cref{tab:reference}, replacing the task-ID conditioning with language-only goal tokens yields 74.2\% overall success, whereas introducing a task-visual branch without FiLM improves the performance to 81.7\% under full-dimensional regularization. This 7.5 percentage-point improvement suggests that language embeddings alone are insufficient to fully capture the visual information required for action generation. Incorporating task-specific visual representations allows the policy to combine language instructions with scene-dependent features, improving its ability to select appropriate actions.

Compared with the task-ID reference, the complete JEPA Dots model further improves overall success from 76.0\% to 84.9\%, with particularly substantial gains on LIBERO-Goal (67.1\% to 88.5\%) and LIBERO-10 (53.1\% to 64.8\%). These results support the effectiveness of incorporating task-aware visual representations into the JEPA-based control framework, although the overall improvement also reflects the contributions of language conditioning and subspace regularization.

\subsection{Effect of language-conditioned FiLM}\label{sec:task_results}
\begin{figure}[t]
\centering\includegraphics[width=\columnwidth]{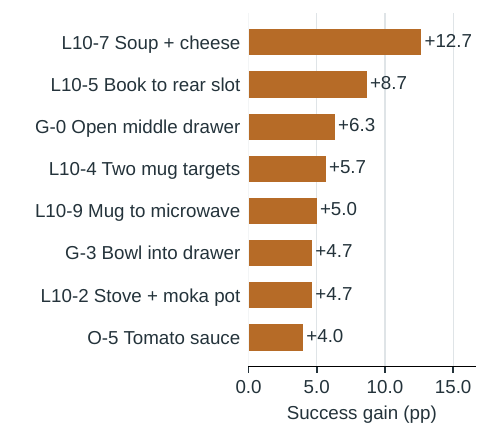}
\caption{\textbf{Selected task-level FiLM gains.} The eight largest positive differences from the matched no-FiLM subspace model, pooled over three runs with 300 episodes per task and model. Complete results for all 40 tasks, including decreases, appear in the supplement.}
\label{fig:task_gains}
\end{figure}

\input{data/replication_table}
We further examine whether explicitly modulating visual features with language improves task-conditioned control. In the reference experiment, adding residual FiLM to the task branch under subspace regularization increases overall success from 83.8\% to 84.9\%, with improvements on LIBERO-Object (94.0\% to 95.8\%) and LIBERO-10 (60.6\% to 64.8\%). As illustrated in \cref{fig:task_gains}, the benefits are particularly pronounced for several manipulation tasks requiring different object interactions and target placements. For instance, success on the LIBERO-10 soup-and-cheese task increases by 12.7 percentage points across three runs.

However, the improvement is not uniform across tasks or training runs. As shown in \cref{tab:replication}, the complete model and its no-FiLM counterpart achieve nearly identical overall success rates across three independent runs (81.9\% and 81.9\%). FiLM therefore contributes primarily through task-dependent feature modulation rather than a consistent improvement in aggregate performance. These findings suggest that language-conditioned visual adaptation can benefit specific manipulation tasks, while its effectiveness depends on the task characteristics.

\subsection{Effect of subspace regularization}
We next investigate the role of subspace Gaussian regularization in learning dynamics representations. As shown in \cref{tab:replication}, replacing full-dimensional regularization with eight-subspace SIGReg improves the three-run average success rate from 80.7\% to 81.9\% within the FiLM architecture. Meanwhile, the mean validation world-prediction MSE decreases from \ensuremath{60.0\times10^{-4}} to \ensuremath{52.6\times10^{-4}}, corresponding to a relative reduction of 12.3\%.

These results suggest that subspace regularization can improve latent prediction quality while benefiting downstream control. By applying Gaussian regularization across multiple low-dimensional projections, the method provides additional constraints on the learned dynamics representations without increasing inference-time computation. Although latent prediction error does not directly measure physical-state accuracy, the consistent MSE reductions across three runs support the usefulness of subspace regularization for jointly learned world--action representations.

\subsection{Verification of task--dynamics separation}
Finally, we examine whether the proposed dual-latent architecture achieves its intended separation between task-dependent control and instruction-independent dynamics prediction. We switch language instructions while keeping the observations, proprioceptive states, supplied actions, model parameters, and action-sampling randomness fixed. The resulting dynamics latents and world predictions exhibit zero maximum absolute change, confirming the instruction-invariant prediction behavior enforced by the masked attention mechanism.

In contrast, the task representations and generated actions respond to instruction changes. With FiLM, task-latent L2 differences range from 3.2 to 6.7, while action-chunk differences range from 1.2 to 4.9. In the same-scene goal-switching evaluation, JEPA Dots successfully executes both requested goals in 50 out of 60 scenes. These results verify that the two latent pathways fulfill their distinct conditioning roles: dynamics representations support action-conditioned world prediction, while task representations enable language-dependent action generation. This separation provides a structured interface for integrating JEPA-based predictive learning with language-guided robotic control.

%% file: data/external_table.tex
\begin{table*}[t]
\centering\small
\caption{\textbf{Comparison with other controllers.} Success rates (\%) and parameters in millions. FM: flow matching. DOTS and V-JEPA totals include frozen MiniLM ($\approx22.7$M); trainable counts exclude it. V-JEPA rows average three rollout streams of one trained checkpoint per size (6,000 episodes); DOTS uses the reference experiment (4,000 episodes). Published policy scores and counts retain their source protocols; -- denotes an unavailable trainable count. Best success rates are shown in bold.}
\label{tab:external}
\setlength{\tabcolsep}{3pt}
\begin{tabular*}{\textwidth}{@{\extracolsep{\fill}}ld{6.2}d{2.2}d{2.2}d{2.2}d{2.2}d{2.2}d{2.2}@{}}\toprule
\multicolumn{1}{l}{Model} & \multicolumn{1}{c}{Total (M)} & \multicolumn{1}{c}{Train. (M)} & \multicolumn{1}{c}{Spatial} & \multicolumn{1}{c}{Object} & \multicolumn{1}{c}{Goal} & \multicolumn{1}{c}{LIBERO-10} & \multicolumn{1}{c}{Overall}\\\midrule
V-JEPA 2.1-B + FM & \approx 115.1 & 5.6 & 72.2 & 85.8 & 80.4 & 36.8 & 68.8\\
V-JEPA 2.1-L + FM & \approx 333.1 & 5.7 & 73.0 & 81.9 & 83.3 & 41.4 & 69.9\\
NanoVLA-S~\cite{chen2025nanovla} & 161.0 & \multicolumn{1}{c}{--} & 81.6 & 93.6 & \mathbf{89}.\mathbf{6} & 49.8 & 78.7\\
SmolVLA-240M~\cite{shukor2025smolvla} & 240.0 & \multicolumn{1}{c}{--} & 87.0 & 93.0 & 88.0 & 63.0 & 82.8\\
JEPA Dots & \approx 39.1 & 16.4 & \mathbf{90}.\mathbf{4} & \mathbf{95}.\mathbf{8} & 88.5 & \mathbf{64}.\mathbf{8} & \mathbf{84}.\mathbf{9}\\
\bottomrule\end{tabular*}
\end{table*}

%% file: data/reference_table.tex
\begin{table*}[t]
\centering\small
\caption{\textbf{Reference experiment.} Success rates (\%) on the four LIBERO suites, with 4,000 episodes per model. Full and subspace denote full-dimensional and eight-subspace SIGReg. Values are rounded to one decimal. Best success rates are shown in bold, including displayed ties.}
\label{tab:reference}
\setlength{\tabcolsep}{3pt}
\begin{tabular*}{\textwidth}{@{\extracolsep{\fill}}ld{2.2}d{2.2}d{2.2}d{2.2}d{2.2}@{}}\toprule
\multicolumn{1}{l}{Variant} & \multicolumn{1}{c}{Spatial} & \multicolumn{1}{c}{Object} & \multicolumn{1}{c}{Goal} & \multicolumn{1}{c}{LIBERO-10} & \multicolumn{1}{c}{Overall}\\\midrule
Task-ID reference & 91.0 & 92.6 & 67.1 & 53.1 & 76.0\\
Language-only goals & 84.5 & 87.8 & 85.4 & 39.0 & 74.2\\
Task head, no FiLM, full & 91.2 & 94.0 & 85.3 & 56.2 & 81.7\\
Task head + FiLM, full & 90.4 & 91.8 & 86.3 & 57.4 & 81.5\\
Task head, no FiLM, subspace & \mathbf{92}.\mathbf{2} & 94.0 & 88.2 & 60.6 & 83.8\\
JEPA Dots (complete) & 90.4 & \mathbf{95}.\mathbf{8} & \mathbf{88}.\mathbf{5} & \mathbf{64}.\mathbf{8} & \mathbf{84}.\mathbf{9}\\
\bottomrule\end{tabular*}
\end{table*}

%% file: data/replication_table.tex
\begin{table*}[t]
\centering\small
\caption{\textbf{Three independent training runs.} Mean success rates (\%) from 12,000 episodes per variant, or 3,000 per suite. Overall entries give the mean and sample SD of the three trained-model success rates. Best means are shown in bold, including displayed ties.}
\label{tab:replication}
\setlength{\tabcolsep}{3pt}
\begin{tabular*}{\textwidth}{@{\extracolsep{\fill}}ld{2.2}d{2.2}d{2.2}d{2.2}d{2.2}@{\extracolsep{0pt}\,$\pm$\,}d{1.2}@{}}\toprule
\multicolumn{1}{l}{Variant} & \multicolumn{1}{c}{Spatial} & \multicolumn{1}{c}{Object} & \multicolumn{1}{c}{Goal} & \multicolumn{1}{c}{LIBERO-10} & \multicolumn{2}{c}{Overall $\pm$ SD}\\\midrule
FiLM, full regularization & 87.8 & 93.2 & 85.1 & 56.7 & 80.7 & 0.8\\
No FiLM, subspace & \mathbf{90}.\mathbf{3} & 93.9 & \mathbf{86}.\mathbf{8} & 56.7 & \mathbf{81}.\mathbf{9} & 1.6\\
JEPA Dots (complete) & 88.3 & \mathbf{94}.\mathbf{1} & \mathbf{86}.\mathbf{8} & \mathbf{58}.\mathbf{4} & \mathbf{81}.\mathbf{9} & 2.6\\
\bottomrule\end{tabular*}
\end{table*}

%% file: sections/06_discussion.tex
\section{Discussion and outlook}
\label{sec:discussion}
\paragraph{What the conditioning separation enforces.}
DOTS gives the two branches different conditioning inputs while training their visual backbone jointly. The dynamics path predicts from history and supplied actions; the task path can change with language. By separating instruction-independent dynamics prediction from language-conditioned task representations, this interface connects JEPA-based predictive learning with flow-matching action generation. The reference experiment and task-level analysis show useful control behavior with this interface. Instruction-switching experiments further confirm unchanged world predictions under fixed observations and supplied actions, while policy outputs respond to language instructions. The repeated-run ablations locate FiLM's benefits more precisely: its effect depends on the task, and understanding that dependence is an important next step. Together with the compact-model comparisons, these findings suggest that the organization of predictive and task-relevant latent information matters for efficient control.

\paragraph{Self-improvement through policy adaptation.}
\label{sec:self_improvement}
Beyond the current imitation-learning setting, the interface also suggests a way to adapt from experience: execute the policy, verify the outcome, combine successful trajectories with demonstrations, and update the task and action adapters. RoboCat and offline-to-online learning use collected experience for later policy updates, while RISE studies improvement through model-generated rollouts~\cite{bousmalis2023robocat,ball2023rlpd,yang2026rise}.

An exploratory two-task pilot collected 40 rollouts, of which 34 succeeded. After 200 adapter-update steps with the dynamics path frozen, the mixture of successful experience and demonstrations achieved 53/60 successes on held-out rollouts. The initial policy achieved 44/60, and an expert-data control with the same update budget achieved 49/60. World MSE remained $51.0\times10^{-4}$ under this protocol. These results motivate a larger study of selective adaptation over repeated interaction cycles.

Keeping the predictor fixed during adaptation requires freezing its entire computation path, including the shared visual encoder. Separating instruction inputs by itself does not isolate parameter updates. Outcome-verification accuracy and held-out evaluation will be central to assessing whether gains accumulate over successive cycles. The supplement describes the pilot separately from the 40-task benchmark.

\paragraph{Evaluation coverage and extensions.}
The experiments cover 40 trained tasks with their canonical instructions. Evaluation on new instructions, object configurations, and real robots would establish how the learned interface transfers beyond this setting. Generalization to these settings remains untested. A matched policy-only control would help isolate the contribution of predictive supervision. Physical-state probes could test which scene variables are retained by the dynamics latent. Deployment currently samples actions directly; using predictions to evaluate candidate actions is a further planning question. Common-hardware latency measurements would add a practical resource measure to the parameter counts reported here.

\section{Conclusion}
\label{sec:conclusion}
We presented \method, a compact dual-latent world--action model for language-guided robot control. It combines LeWM-style end-to-end predictive learning with masked world--action coupling drawn from LeWAM. Language conditions task perception and action generation, while dynamics prediction reads visual history and supplied actions. The control network has 16.4M parameters, or approximately 39.1M including the frozen sentence encoder, and achieves 84.9\% LIBERO success in the reference experiment, with Goal success increasing from 67.1\% to 88.5\%. Repeated-run analyses show task-specific FiLM gains and lower latent-prediction MSE with subspace regularization. Together with the instruction-switching analysis, these results support structured task--dynamics separation for compact control and offer a concrete setting for studying selective policy adaptation.

%% file: sections/07_llm_statement.tex
\section*{LLM Usage Statement}
We used OpenAI ChatGPT to search and review the literature, discuss experimental protocols, and improve the manuscript’s language and clarity. OpenAI Codex helped with code generation, architecture diagrams, and plotting scripts.

%% file: sections/supplementary_body.tex
\section{DOTS training and deployment}
This supplement reports training settings, complete task counts, checkpoint statistics, and additional studies. Runs A--C identify independently trained DOTS ablations. The V-JEPA comparisons use one trained checkpoint per size and separate evaluation streams, as detailed in Section~\ref{sec:vjepa_protocol}. Run A is the reference experiment: its full model achieves 84.9\% success. The three-run mean is 81.9\%. The tables distinguish these two reporting scopes. Fractional table entries are rounded to one decimal; integer counts are unchanged. Scale factors preserve small losses and exact preprocessing or optimizer settings. Unrounded experiment records are retained in the source data.

\begin{table}[ht]\centering\small
\caption{Architecture and preprocessing.}\label{tab:settings1}
\begin{tabular}{@{}p{.32\linewidth}p{.63\linewidth}@{}}\toprule Setting & Configuration\\\midrule
Views and history & External and wrist RGB; two observed frames\\
Capture and resize & $128\times128$ uint8; HWC to CHW; divide by 255; bilinear resize to $224\times224$, \texttt{align\_corners=False}\\
Image normalization & ImageNet mean $(485.0,456.0,406.0)\times10^{-3}$ and SD $(229.0,224.0,225.0)\times10^{-3}$\\
Encoder crop & $202\times202$ random crop during training; center crop at evaluation\\
Backbone & Shared ResNet-18, jointly trained from random initialization\\
Latents & Separate 192D dynamics and task heads for each view\\
Language & Frozen \texttt{all-MiniLM-L6-v2}; 384D output projected to 192D\\
Task conditioning & Zero-initialized residual FiLM and two view-specific goal tokens\\
Proprioception & 15D joint, gripper, end-effector position, and orientation fields\\
State and action scaling & Per-coordinate training mean/SD, with SD floor $1.0\times10^{-6}$\\
Actions & 7D; eight-step chunk; masked zero padding at episode boundaries\\
MoT & Four layers, four heads, width 192\\
World target & Future dynamics latent at $t+2$\\
Regularization & Full-dimensional SIGReg or eight fixed 24D subspaces; 1,024 sketches total\\
Subspace matching & Corresponding FiLM and no-FiLM runs reuse identical bases\\\bottomrule
\end{tabular}\end{table}

\begin{table}[ht]\centering\small
\caption{Optimization and closed-loop evaluation.}\label{tab:settings2}
\begin{tabular}{@{}p{.32\linewidth}p{.63\linewidth}@{}}\toprule Setting & Recorded value\\\midrule
Training hardware & One NVIDIA A100 per model process\\
Parallel experiments & Independent model processes scheduled across GPUs\\
Training duration & Not reported\\
Epochs and minibatches & 60 epochs, 500 minibatches per epoch; 30,000 forward/backward minibatches\\
Batch and accumulation & 32 per minibatch; accumulation 2; effective batch approximately 64\\
Optimizer updates & Approximately 15,000 per completed run\\
Optimizer & AdamW; learning rate $1.0\times10^{-4}$; weight decay $1.0\times10^{-4}$; default PyTorch betas and epsilon\\
Schedule & Five-epoch linear warmup, then 55-epoch cosine decay to zero\\
Precision and clipping & bf16 autocast; gradient norm clipped at 1.0\\
Loss weights & Action 1.0; world 1.0; dynamics regularization $4.0\times10^{-2}$\\
Selection & Minimum validation action flow-matching loss\\
Trajectory split & 10.0\% held-out demonstrations; seeded hash; split may vary across runs\\
Validation windows & 960 per run; reference run has 301,593 training windows\\
Evaluation & 40 trained tasks; 50 official initial states each; two simulation random streams\\
Episode counts & 4,000 per model per run; 12,000 across three runs\\
Episode horizon & At most 600 environment steps\\
Inference & Eight ODE steps; execute two of eight actions; unnormalize and clip to $[-1,1]$\\
World use at inference & No candidate scoring or simulator look-ahead in the main controller\\\bottomrule
\end{tabular}\end{table}

\clearpage\section{Run counts, checkpoints, and model size}
\begin{table}[ht]\centering\small
\caption{Exact closed-loop successes. Each run has 4,000 episodes per model. Bold marks the largest count in each column.}\label{tab:run_counts}
\begin{tabular}{@{}lrrrr@{}}\toprule Variant & Run A & Run B & Run C & Total\\\midrule
FiLM, full regularization &3259&3227&3198&9684\\
No FiLM, subspace &3350&\textbf{3239}&\textbf{3242}&\textbf{9831}\\
JEPA Dots &\textbf{3395}&3233&3198&9826\\\bottomrule\end{tabular}\end{table}

\begin{table}[ht]\centering\small
\caption{Selected checkpoints. Epoch numbering is one-based. World MSE is reported at the checkpoint selected by action loss. Values multiply the power of ten in each heading and are rounded to one decimal. Bold marks the lowest displayed loss within each run; latent MSE uses each model's learned coordinates.}\label{tab:checkpoints}
\begin{tabular}{@{}llrd{3.2}d{3.2}@{}}\toprule Run & Variant & Epoch/60 & \multicolumn{1}{c}{Action loss ($\times10^{-3}$)} & \multicolumn{1}{c}{World MSE ($\times10^{-4}$)}\\\midrule
A & Language-only goals & 55 & 469.4 & 110.6\\
A & No FiLM, full & 57 & \mathbf{466}.\mathbf{0} & 53.0\\
A & FiLM, full & 56 & 468.7 & 53.9\\
A & No FiLM, subspace & 60 & 482.2 & 45.1\\
A & JEPA Dots & 60 & 485.2 & \mathbf{44}.\mathbf{5}\\
A & Dense task head & 58 & 481.2 & 89.9\\
B & FiLM, full & 60 & 517.7 & 65.4\\
B & No FiLM, subspace & 56 & 517.6 & \mathbf{62}.\mathbf{5}\\
B & JEPA Dots & 60 & \mathbf{517}.\mathbf{3} & 63.5\\
C & FiLM, full & 57 & \mathbf{466}.\mathbf{1} & 60.7\\
C & No FiLM, subspace & 60 & 483.5 & 50.0\\
C & JEPA Dots & 60 & 482.3 & \mathbf{49}.\mathbf{7}\\
\bottomrule\end{tabular}\end{table}

\paragraph{Loss conventions and run variation.}
Learning rates are logged after the epoch-end scheduler update. Run A's selected full-model checkpoint stores epoch 59 using zero-based indexing, corresponding to epoch 60 in the table. Checkpoints are selected by validation action loss. Since trajectory splits can vary between runs, their action losses refer to different validation sets. Matched variants use the same split rule within each run.

\paragraph{Parameter accounting.}
The full control network has 16,437,605 parameters. The no-FiLM subspace, language-only, and dense task variants have 16,239,973, 16,087,045, and 16,568,645 parameters, respectively. These counts exclude frozen MiniLM. FiLM adds 197,632 parameters, or approximately 1.2\% relative to the matched no-FiLM model. The sentence encoder contributes approximately 22.7M, bringing the total to about 39.1M. The rounded total is estimated from the component configurations. Caching text features saves repeated encoding work while leaving the parameter count unchanged.

\paragraph{Statistical interpretation.}
The mean FiLM effect over three runs is $0.0$ pp; the individual effects are $1.1$, $-0.2$, and $-1.1$ pp. Replacing full-dimensional regularization with the subspace objective within FiLM gives $3.4$, $0.2$, and $0.0$ pp. Standard deviations are sample SDs of the three trained-model success rates. Estimating paired confidence intervals or failure-to-success transitions would require episode-level outcome pairs.

\clearpage\section{All 40 task results}
\begin{figure}[ht]\centering
\includegraphics[width=\textwidth]{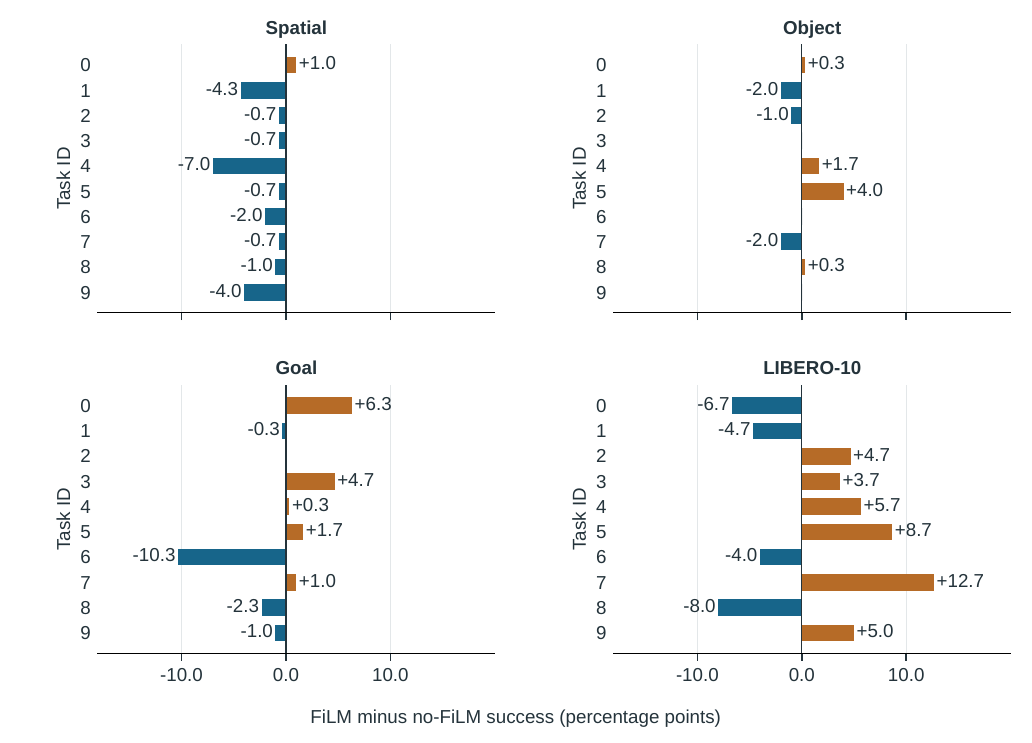}
\caption{All 40 task differences across three training runs. Orange marks positive FiLM effects, blue negative effects. Each difference uses 300 episodes per model and task.}
\label{fig:all_tasks}\end{figure}

The main paper selects the eight largest positive FiLM-minus-no-FiLM differences in the three-run aggregate, breaking ties by suite and local task ID. Across all tasks, sixteen improve, twenty decrease, and four are unchanged. Figure~\ref{fig:all_tasks} shows the distribution, and Table~\ref{tab:alltasks} lists every count. These complete results give the context for the positive examples.

\clearpage
\begingroup\small\setlength{\tabcolsep}{5pt}\input{data/all_tasks_table}\endgroup

\clearpage\section{Instruction switching and task-head variants}
\paragraph{Fixed-input intervention.}
We switch between Object instructions involving alphabet soup and tomato sauce, and Goal instructions placing a bowl on a stove or plate. The checkpoint, observed state, supplied world actions, and action-sampling randomness are held fixed. Across three runs, dynamics latents and world predictions have zero maximum absolute difference. For DOTS, task-latent L2 differences are approximately 3.2--6.7 and action-chunk differences 1.2--4.9. The no-FiLM task latent stays fixed, but its actions can change through the language goal tokens.

\begin{table}[ht]\centering
\caption{Same-scene instruction switching. Each of 60 scenes is evaluated separately with two goals. ``Both'' counts scenes where both requested rollouts succeed. Bold marks the largest count in each column.}
\begin{tabular}{@{}lrrr@{}}\toprule Variant & Bowl to stove & Bowl to plate & Both\\\midrule
No FiLM, subspace &\textbf{56/60}&\textbf{60/60}&\textbf{56/60}\\
JEPA Dots &55/60&53/60&50/60\\\bottomrule\end{tabular}\end{table}

\paragraph{Feature-map intervention.}
Removing FiLM changes actions and spatial-softmax keypoints in the tested scenes. Object-mask analyses show no consistent grounding improvement. A spatially constant channel shift is largely canceled by spatial softmax, while scale modulation accounts for the observed keypoint changes. These probes distinguish sensitivity to language from correct object selection.

\paragraph{Dense task-vision variant.}
Replacing task pooling with language-query attention over $4\times4$ spatial tokens per view gives 3,049/4,000 successes (76.2\%) in the reference experiment. DOTS gives 3,395/4,000 (84.9\%). The dense variant's suite rates are 84.6, 87.0, 81.2, and 52.1\% in Spatial, Object, Goal, and LIBERO-10 order.

\section{Transfer, ranking, and adaptation studies}
The following studies use different initializations, trainable components, or inference procedures. Their results are reported separately from the end-to-end DOTS benchmark.

\paragraph{Frozen transfer pilot.}
A previous task-ID model, evaluated on Object and Goal with its ID input removed, succeeds in 584/2,000 episodes (29.2\%). Adding language goal tokens gives 1,268/2,000 (63.4\%); adding both a task head and FiLM gives 1,714/2,000 (85.7\%). The latter comparison changes both components. The frozen world path retains MSE \ensuremath{87.4\times10^{-4}}. The initialization has already seen all 40 tasks.

\paragraph{Separate LeWM+MoT ranking study.}
This study trains a ViT-based LeWM first, freezes it, and then trains the policy. Language queries retrieve goal images from training demonstrations. Four inference modes are each evaluated over 2,400 episodes across three runs. Direct flow sampling and random ranking each give 1,409 successes (58.7\%). An action-shuffle control gives 1,414 (58.9\%), and selecting among eight candidates by terminal predicted latent goal distance gives 1,211 (50.5\%). The lower success of distance-based ranking shows the importance of evaluating a latent goal metric in closed-loop control.

\paragraph{Small adaptation pilot.}
The two-task pilot collects 40 rollouts and retains 34 successful trajectories. Task and action adapters are updated for 200 steps with the dynamics path frozen. Each condition is evaluated on 60 held-out rollouts: the initial policy succeeds 44 times, the expert-data control with the same update budget 49 times, and the mixture of successful experience and demonstrations 53 times. World MSE stays at \ensuremath{51.0\times10^{-4}} under this frozen-path protocol. The pilot uses a separate protocol from the 40-task benchmark and examines a single adaptation cycle.

\input{sections/supp_vjepa_protocol}

%% file: data/all_tasks_table.tex
\begin{longtable}{@{}llrrrd{3.2}@{}}
\caption{Complete task counts. Each model has 300 episodes per task. Descriptions are shortened; IDs are local to each suite. Bold marks the largest success count within each task, including ties. Differences are rounded to one decimal.}\label{tab:alltasks}\\
\toprule Task & Description & No FiLM & DOTS & FiLM full & \multicolumn{1}{c}{$\Delta$ pp}\\\midrule\endfirsthead
\toprule Task & Description & No FiLM & DOTS & FiLM full & \multicolumn{1}{c}{$\Delta$ pp}\\\midrule\endhead
S0 & Bowl between plate and ramekin to plate & 293 & \textbf{296} & 291 & +1.0\\
S1 & Bowl beside ramekin to plate & \textbf{237} & 224 & 172 & -4.3\\
S2 & Bowl at table center to plate & \textbf{297} & 295 & 296 & -0.7\\
S3 & Bowl on cookie box to plate & 287 & 285 & \textbf{292} & -0.7\\
S4 & Bowl in top drawer to plate & 247 & 226 & \textbf{250} & -7.0\\
S5 & Bowl on ramekin to plate & 254 & 252 & \textbf{255} & -0.7\\
S6 & Bowl beside cookie box to plate & \textbf{298} & 292 & 294 & -2.0\\
S7 & Bowl on stove to plate & \textbf{280} & 278 & \textbf{280} & -0.7\\
S8 & Bowl beside plate to plate & 243 & 240 & \textbf{253} & -1.0\\
S9 & Bowl on cabinet to plate & \textbf{272} & 260 & 251 & -4.0\\
O0 & Alphabet soup to basket & 290 & \textbf{291} & 276 & +0.3\\
O1 & Cream cheese to basket & 231 & 225 & \textbf{239} & -2.0\\
O2 & Salad dressing to basket & \textbf{298} & 295 & 290 & -1.0\\
O3 & BBQ sauce to basket & \textbf{285} & \textbf{285} & 284 & +0.0\\
O4 & Ketchup to basket & 295 & \textbf{300} & 289 & +1.7\\
O5 & Tomato sauce to basket & 263 & \textbf{275} & 259 & +4.0\\
O6 & Butter to basket & 277 & 277 & \textbf{285} & +0.0\\
O7 & Milk to basket & \textbf{296} & 290 & 290 & -2.0\\
O8 & Chocolate pudding to basket & 289 & \textbf{290} & 289 & +0.3\\
O9 & Orange juice to basket & \textbf{294} & \textbf{294} & \textbf{294} & +0.0\\
G0 & Open middle drawer & 258 & \textbf{277} & 264 & +6.3\\
G1 & Bowl to stove & 284 & 283 & \textbf{291} & -0.3\\
G2 & Wine bottle to cabinet top & \textbf{287} & \textbf{287} & 263 & +0.0\\
G3 & Open top drawer and insert bowl & 156 & \textbf{170} & 142 & +4.7\\
G4 & Bowl to cabinet top & 292 & \textbf{293} & \textbf{293} & +0.3\\
G5 & Push plate in front of stove & 254 & 259 & \textbf{262} & +1.7\\
G6 & Cream cheese to bowl & \textbf{257} & 226 & 244 & -10.3\\
G7 & Turn on stove & 297 & \textbf{300} & 299 & +1.0\\
G8 & Bowl to plate & 284 & 277 & \textbf{285} & -2.3\\
G9 & Wine bottle to rack & \textbf{234} & 231 & 211 & -1.0\\
L0 & Soup and tomato sauce to basket & \textbf{134} & 114 & 103 & -6.7\\
L1 & Cream cheese and butter to basket & 226 & 212 & \textbf{247} & -4.7\\
L2 & Turn on stove and place moka pot & 212 & \textbf{226} & 207 & +4.7\\
L3 & Bowl into bottom drawer and close & 252 & \textbf{263} & 258 & +3.7\\
L4 & Two mugs to separate plates & 130 & \textbf{147} & 144 & +5.7\\
L5 & Book into rear compartment & 238 & \textbf{264} & 253 & +8.7\\
L6 & Mug to plate and pudding beside it & \textbf{57} & 45 & 37 & -4.0\\
L7 & Soup and cream cheese to basket & 205 & \textbf{243} & 221 & +12.7\\
L8 & Two moka pots to stove & \textbf{120} & 96 & 95 & -8.0\\
L9 & Mug into microwave and close & 128 & \textbf{143} & 136 & +5.0\\
\bottomrule\end{longtable}

%% file: sections/supp_vjepa_protocol.tex
\clearpage
\section{V-JEPA 2.1 baseline configuration}
\label{sec:vjepa_protocol}
The V-JEPA 2.1-B and -L comparisons train action policies on frozen pretrained visual features. They use one training run per encoder size, followed by three rollout streams for each selected checkpoint. The B/L comparison holds the demonstration assignments, flow-policy architecture, optimization budget, and evaluation procedure fixed. Relative to DOTS, the setup also changes the encoder, its freezing, the training objective, image preprocessing, and evaluation count. These comparisons therefore concern complete learning setups rather than an isolated world-loss ablation.

\subsection{Frozen encoders and parameter accounting}
We load only the \texttt{ema\_encoder} weights from the official distilled checkpoints \texttt{vjepa2\_1\_vitb\_dist\_vitG\_384.pt} and \texttt{vjepa2\_1\_vitl\_dist\_vitG\_384.pt}, with strict key and shape matching. The encoders remain frozen and in evaluation mode; feature extraction uses no gradients. Pretrained predictors and optimizer states are not loaded. Although the checkpoint reference resolution is $384\times384$, each image is encoded independently at $224\times224$ through the image-tokenizer path, using the official interpolated rotary positional encoding. Each frame produces 196 patch tokens. Future frames are never policy inputs.

\begin{table}[ht]\centering\small
\caption{V-JEPA comparison components. Exact counts are shown except for MiniLM and the resulting approximate full totals. The original vision--policy totals exclude the separate frozen sentence encoder; the main table includes it consistently for DOTS and both baselines. Parameter counts are not performance rankings.}
\label{tab:vjepa_parameters}
\setlength{\tabcolsep}{6pt}
\begin{tabular*}{\textwidth}{@{\extracolsep{\fill}}lrr@{}}\toprule
Component or configuration & V-JEPA 2.1-B + FM & V-JEPA 2.1-L + FM\\\midrule
Encoder depth / width / heads & 12 / 768 / 12 & 24 / 1,024 / 16\\
Frozen visual encoder & 86,833,152 & 304,680,960\\
Trainable flow module & 4,596,871 & 4,596,871\\
Other trainable adapters and conditioning & 969,918 & 1,150,142\\
Total trainable & 5,566,789 & 5,747,013\\
Vision--policy subtotal & 92,399,941 & 310,427,973\\
Separate frozen MiniLM & $\approx22.7$M & $\approx22.7$M\\
All-component total & $\approx115.1$M & $\approx333.1$M\\\bottomrule
\end{tabular*}
\end{table}

The separate frozen \texttt{all-MiniLM-L6-v2} encoder produces 384D instruction embeddings, cached for all 40 tasks. Caching avoids repeated text encoding without changing the component count. DOTS has 16,437,605 trainable visual and control parameters, or approximately 39.1M with the same sentence encoder. Thus, the smaller DOTS total should be distinguished from its larger trainable count relative to these frozen-encoder policies.

\subsection{Adapters and action-only flow policy}
All adapters and flow-policy parameters are initialized from scratch; no trained DOTS policy weights are transferred. FiLM, action AdaLN, and the final proprioception projection use zero initialization. The history adapter mean-pools patch tokens, projects the encoder width to 192D, and adds a projected 15D proprioceptive state. The task adapter projects patch features to 512 channels, applies language FiLM, and uses a 32-keypoint spatial softmax followed by a $64\rightarrow384\rightarrow192$ MLP. Each latest per-view task latent is concatenated with the projected 192D language embedding to form a 384D goal before token embedding.

The policy separately implements the action-reachable MoT subgraph, retaining the history/goal state-stream projections. It has four transformer blocks, width 192, four heads, feed-forward width 768, and dropout 0.1. The sequence contains four history tokens, two goal tokens, and eight noisy-action tokens, each embedded to 192D. Actions attend to history, goals, and current or preceding noisy actions. A sinusoidal flow-time embedding and AdaLN-Zero condition the action tokens. World-state queries and clean-action slots are omitted because the action branch cannot attend to them under the mask.

The sole objective is rectified-flow velocity MSE. For standardized expert actions $A$, draw independent $\epsilon\sim\mathcal N(0,I)$ and $s\sim\mathcal U(0,1)$, use $(1-s)\epsilon+sA$ as input, and regress $A-\epsilon$. Padded positions are masked. Neither world-prediction loss nor SIGReg is applied. Each policy outputs eight 7D actions.

\clearpage\subsection{Demonstrations, preprocessing, and training}
\begingroup\small
\begin{longtable}{@{}p{.30\textwidth}p{.65\textwidth}@{}}
\caption{Data and optimization for both V-JEPA sizes. An epoch denotes a fixed sampling budget, not a complete traversal of every available window.}\label{tab:vjepa_training}\\
\toprule Setting & Shared configuration\\\midrule\endfirsthead
\toprule Setting & Shared configuration (continued)\\\midrule\endhead
Suites and demonstrations & Ten tasks each from Spatial, Object, Goal, and LIBERO-10; 50 demonstrations per task, 2,000 total.\\
Training / validation & 45 / 5 complete demonstrations per task, or 1,800 / 200 total. No offline test split; final outcomes use simulator rollouts.\\
Deterministic split & Rank demonstrations by SHA-256 of the split identifier, task key, and demonstration name; hold out the first five per task. Both sizes use identical assignments. The split identifier is independent of the training initialization.\\
Window sampling & Task-balanced weighted sampling with replacement.\\
Cameras and history & External \texttt{agentview\_rgb} and wrist \texttt{eye\_in\_hand\_rgb}, at $t-1$ and $t$.\\
Image preparation & Native $128\times128$ RGB, bilinear resize to 224, $202\times202$ crop, then resize to $224\times224$ and ImageNet normalization. Random crops in training; center crops in validation and simulator evaluation.\\
State and action scaling & 15D proprioception and 7D actions use the same saved standardization statistics as DOTS.\\
Sampling budget & 60 epochs, 500 minibatches per epoch, 32 windows per minibatch; 30,000 minibatches and 960,000 window draws including repeats.\\
Gradient accumulation & Two minibatches; effective batch size 64, 250 optimizer updates per epoch, 15,000 in total.\\
Optimizer & AdamW on trainable parameters only; learning rate and weight decay $1.0\times10^{-4}$, betas $(0.9,1-1.0\times10^{-3})$, epsilon $1.0\times10^{-8}$.\\
Schedule and numerics & Five-epoch linear warmup followed by cosine decay; BF16 autocast; maximum gradient norm 1.0; eight data-loader workers.\\
Validation & After every epoch, 24 fixed windows per task, 960 total. The same flow-noise/time stream is reset for each validation pass.\\
Checkpoint selection & Save the checkpoint with the lowest mean validation action-flow loss. Simulator success is not used for selection.\\\bottomrule
\end{longtable}
\endgroup
The selected B checkpoint is epoch 56 with validation loss \ensuremath{468.7\times10^{-3}}; the selected L checkpoint is epoch 57 with validation loss \ensuremath{472.3\times10^{-3}}. Epochs are one-based. These values describe model selection, not repeated-training estimates.

\subsection{Closed-loop evaluation and aggregation}
\begingroup\small
\begin{longtable}{@{}p{.30\textwidth}p{.65\textwidth}@{}}
\caption{V-JEPA evaluation protocol. All three rollout sets use the same selected weights; they are not independent training runs.}\label{tab:vjepa_evaluation}\\
\toprule Setting & Configuration\\\midrule\endfirsthead
\toprule Setting & Configuration (continued)\\\midrule\endhead
Rollout count & Three evaluation streams, 50 episodes per task per stream: 150 per task, 1,500 per suite, and 6,000 per model; 12,000 across both sizes.\\
Initial states & Official indices 0--49 per task, reused across streams.\\
Episode randomization & Each episode uses the base stream identifier plus the suite-local task index multiplied by 1,000, plus the episode index.\\
Step budget & Ten warmup steps, then at most 600 policy-controlled environment steps; terminate earlier on success.\\
Action sampling & Independent Gaussian action noise and eight fixed Euler integration steps.\\
Execution & Predict eight actions, inverse-standardize, clip to $[-1,1]$, and execute the first two before replanning.\\
Success & LIBERO simulator \texttt{env.check\_success()}.\\
Aggregation & Compute success for each evaluation stream, then average the three equal-size rollout sets. Each suite has ten equally weighted tasks.\\\bottomrule
\end{longtable}
\endgroup
\clearpage
The reported Spatial, Object, Goal, and LIBERO-10 rates are (72.2, 85.8, 80.4, 36.8)\% for B and (73.0, 81.9, 83.3, 41.4)\% for L, with reported overall rates of 68.8\% and 69.9\%. We retain the precision of these recorded aggregates and do not infer per-stream uncertainty or exact success counts from rounded percentages. The separate DOTS three-training-run estimates remain in the main paper.

\section{Published comparison sources}
NanoVLA-S scores are taken from Table 1 of its source paper, and SmolVLA-240M scores from Table 2 of its source. Their reported parameter counts are 161.0M total parameters and 240.0M model parameters, respectively. These published evaluations use their original data and protocols; the selected configurations are not an exhaustive leaderboard. SmolVLA reports ten trials per task, compared with 100 per trained DOTS model and 150 per V-JEPA model here.

The complete task tables preserve aggregate outcomes for all DOTS variants. Paired confidence intervals and failure-to-success transitions would require episode-level outcome pairs. Resource comparisons report parameter counts; common-hardware latency and elapsed training time have not been measured here.